\documentclass{article}

\title{Abduction, ASP and Open Logic Programs}

\author{ {\bf P.\ A.\ Bonatti}\\
	   Dipartimento di Tecnologie dell'Informazione\\
	   Universit\`a di Milano\\
	   bonatti@dti.unimi.it
}

\usepackage{proceedings}
\usepackage{latexsym}

\newcommand{\qed}{{\hspace*{\fill} \rule{1.2mm}{2.2mm}}}
\newcommand{\ent}[1]{\models^{#1}}
\newcommand{\crd} {\ent{c }}
\newcommand{\skp} {\ent{s }}
\newcommand{\mcs}{\ent{cs}}
\newcommand{\msc}{\ent{sc}}

\newcommand{\comp}{\mathsf{Comp}}

\newcommand{\tup}[1]{\left\langle \, #1 \, \right\rangle}
\newcommand{\hide}[1]{}

\newcommand{\abducibles}{\ensuremath{\mathsf{Abducibles}}}

\newtheorem{theorem}{Theorem}[section]
\newtheorem{lemma}[theorem]{Lemma}
\newtheorem{corollary}[theorem]{Corollary}
\newtheorem{proposition}[theorem]{Proposition}

\newenvironment{proof}
{\begin{trivlist} \item[] {\bf Proof.}}%
{\qed \end{trivlist}}

\newenvironment{definition}
{\begin{defi} \rm}{\qed \end{defi}}

\newenvironment{example}
{\begin{exa} \rm}{\qed \end{exa}}

\newenvironment{remark}
{\begin{rem} \rm}{\end{rem}}

\newenvironment{theorem*}[2]%
{\begin{trivlist} \item[] {\bf #1~\protect{\ref{#2}}}\it}{\end{trivlist}}

\newtheorem{defi}[theorem]{Definition}
\newtheorem{exa}[theorem]{Example} 
\newtheorem{rem}[theorem]{Remark}

\begin{document} 

	\maketitle

\begin{abstract}
Open logic programs and open entailment have been recently proposed as
an abstract framework for the verification of incomplete
specifications based upon normal logic programs and the stable model
semantics.  There are obvious analogies between open predicates and
abducible predicates.  Their extension is not specified in the
program.  However, despite superficial similarities, there are
features of open programs that have no immediate counterpart in the
framework of abduction and viceversa.  Similarly, open programs cannot
be immediately simulated with answer set programming (ASP).

In this paper we start a thorough investigation of the relationships
between open inference, abduction and ASP.  We shall prove that open
programs generalize the other two frameworks.  Similarities and
differences between the three frameworks will be analyzed formally.
The generalized framework suggests interesting extensions of abduction
under the generalized stable model semantics.  In some cases, we will
be able to reduce open inference to abduction and ASP, thereby
characterizing the computational complexity of credulous and skeptical
open inference for finite, function-free open programs.  At the same
time, the aforementioned reduction opens the way to new applications
of abduction and ASP.

\end{abstract}

	\section{Introduction}

Open logic programs and open entailment \cite{Bo-lpnmr01} have been
recently proposed as an abstract framework for the verification of
incomplete specifications based upon normal logic programs and the
stable model semantics.

An important example of incomplete specifications is given by compound
security policies \cite{BdCS00}, where details such as the set of users
and the formulation of certain subpolicies are tipically unknown at
verification time.  In this setting, it is interesting to verify
whether the policy will necessarily satisfy privacy laws, all parties'
requirements, etc.

Logic-based agents are a second important example.  In IMPACT
\cite{impact}, agent programs must satisfy a property called
\emph{conflict freedom}.  This property depends on a variable state
that extends the agent program.  The state is unknown at verification
time, and can be regarded as a runtime extension of the agent
program.  Conflict freedom should hold for all possible such
extensions.

Incomplete specifications are modelled by identifying a set of
predicates, called \emph{open predicates}, that are not completely
defined in the program.  Different forms of entailment capture what
could possibly be true and what must necessarily be true, across a
space of possible complete specifications of the open predicates.

There are obvious analogies between open predicates and abducible
predicates.  Their extension is not specified in the program.  In the
framework of abduction, a suitable definition of abducible predicates
must be found as part of the reasoning task, while in the framework of
open programs one may also quantify over all possible definitions of
the open predicates (skeptical open inference).  Moreover, abduction
looks for simple definitions of the abducible predicates, consisting
of ground facts, while the framework of open programs admits arbitrary
definitions.  It should be pointed out that arbitrary rules can be
abduced by introducing new abducibles, but the point is whether the
resulting abduction framework must necessarily be infinite when the
set of abducible rules is.

Thus, despite superficial similarities, there are features of open
programs that cannot be immediately reproduced in the framework of
abduction and viceversa.  Similarly, answer set programming (ASP)
cannot trivially simulate open programs, because of the complex
structure of some forms of open inference (called \emph{mixed}
inference), and because of the simultaneous treatment of diverse
Herbrand domains, featured by open program semantics.

In this paper we start a thorough investigation of the relationships
between open inference, abduction and ASP.  We shall consider the
abduction frameworks originated by \cite{EK89}, refined in
\cite{KM90}, and further studied -- from a procedural standpoint -- in
\cite{SI92}.  The semantics is based upon \emph{generalized stable
models}, due to \cite{KM90}.  The main contributions of the paper are
the following.

\begin{itemize}
\item
We shall prove that open programs  generalize
both ASP and the generalized stable model semantics.  A related
benefit is that the applicability range of abduction and ASP is
potentially extended to the intended applications of open programs
(under some restrictions).  As a second benefit, the relationships
between the three frameworks open the way to a cross-fertilization of
the proof procedures for abduction, ASP and open inference.  Moreover,
the generalized framework suggests some interesting extensions:
\begin{itemize}
\item
The generalized stable model semantics does not allow to abduce the
existence of new individuals.  Open logic programs provide such
extension, previously supported only under 3-valued completion
semantics \cite{DDS98} and stratified negation \cite{Sh89}.
Furthermore, open logic programs can express upper bounds on domain
cardinality.  None of the existing approaches handles such
constraints.

\item
The existing abductive procedures for the generalized stable model
semantics do not support nonground inference.  Open inference is a
first step toward nonground abduction under the generalized stable
model semantics.  Moreover, open inference suggests new
computation strategies, e.g., based on lemma generation.

\item
The \emph{mixed} inferences introduced in the open framework suggest a
new form of abduction, that stands to standard abduction as
\emph{diagnosis as entailment} stands to \emph{diagnosis as
satisfaction}.
\end{itemize}

\item
In some cases, we will be able to reduce open inference to ASP.  As a
result, we characterize the complexity of two forms of open inference
for finite and function-free open programs.  Moreover, we show how ASP
can be used to compute open inference in this special case.  A
complete embedding will be left as an open issue.

\item
The (partial) embedding of open programs into ASP makes it possible to
use ASP engines such as Smodels \cite{NS97} for abductive reasoning
under the open version of the generalized stable model semantics.
\end{itemize}

The basic definitions concerning abduction frameworks and open
programs will be recalled in Section~\ref{sec:prel}.  Then we will
study the relationships between open programs and abduction frameworks
in Section~\ref{sec:op-vs-abd}.  In Section \ref{sec:op-vs-asp},
credulous and skeptical open inferences are embedded into standard
credulous and skeptical inferences in ASP.  The main results are
summarized and discussed in Section~\ref{sec:conc}.

	\section{Preliminaries}
	\label{sec:prel}

We assume the reader to be familiar with normal logic programs and the
stable model semantics \cite{GL88}.  We say that a normal logic
program is \emph{consistent} if it has at least one stable model.

An \emph{abduction framework} is a pair $\tup{T,A}$, where $T$ is a
normal logic program and $A$ is a set of \emph{abducible predicates}.
Let $\abducibles(T,A)$ be the set of all ground atoms
$p(t_1,\ldots,t_n)$ such that $p\in A$ and $t_i$ belongs to the
Herbrand domain of $T$ $(1\leq i\leq n)$.

\begin{definition}				\label{def:gen-st-mod}
A \emph{generalized stable model} of an abduction framework
$\tup{T,A}$ is a stable model of $T\cup E$, for some
$E\subseteq\abducibles(T,A)$.
\end{definition}

An \emph{open program} is a triple $\tup{P,F,O}$ where $P$ is a
normal logic program, $F$ is a set of function and constant symbols
not occurring in $P$, and $O$ is a set of predicate symbols.  A
\emph{completion}%
\footnote{
	Note that this notion of completion has nothing to do with the
	standard notion of completion derived from Clark's work.  We
	have not been able to identify an alternative term to express
	correctly the idea of complete predicate specification.
}
 of $\tup{P,F,O}$ is a normal logic program $P'$ such
that 
\begin{enumerate}
\item $P'\supseteq P$,
\item the constant and function symbols of $P'$ occur in $P$ or $F$,
\item if $r \in P'\setminus P$, then $head(r)\in O$.
\end{enumerate}
The set of all the completions of $\tup{P,F,O}$ is denoted by
$\comp(P,F,O)$.  There exist four kinds of \emph{open inference}:
\begin{enumerate}

\item (Credulous open inference) $\tup{P,F,O} \crd \Psi$ 
iff for some $P'\in\comp({P,F,O})$, $P'$ credulously entails $\Psi$.

\item (Skeptical open inference) $\tup{P,F,O} \skp \Psi$
iff for all $P'\in\comp({P,F,O})$, $P'$ skeptically entails $\Psi$.

\item (Mixed open inference I) $\tup{P,F,O} \mcs \Psi$ iff for some
consistent $P'\in\comp({P,F,O})$, $P'$ skeptically entails $\Psi$.

\item (Mixed open inference II) $\tup{P,F,O} \msc \Psi$ iff for each
consistent $P'\in\comp({P,F,O})$, $P'$ credulously entails $\Psi$.
\end{enumerate}

The four open entailments are pairwise dual and form a diamond-shaped
lattice \cite{Bo-lpnmr01}.

\begin{proposition}(Duality)
For all open programs $\Omega$ and all sentences $\Psi$,
\begin{enumerate}
\item $\Omega \crd \Psi$ iff $\Omega \not\skp \neg \Psi$;\medskip
\item $\Omega \msc \Psi$ iff $\Omega \not\mcs \neg \Psi$.
\end{enumerate}
\end{proposition}

\begin{proposition}(Entailment lattice)
Suppose there exists a consistent $P'\in\comp(\Omega)$.  Then, for all
sentences $\Psi$,
\begin{enumerate}
\item $\Omega\skp\Psi$ implies $\Omega\mcs\Psi$	and $\Omega\msc\Psi$;
\item $\Omega\mcs\Psi$ implies $\Omega\crd\Psi$;
\item $\Omega\msc\Psi$ implies $\Omega\crd\Psi$.
\end{enumerate}
\end{proposition}

	\section{Open programs and abduction}
	\label{sec:op-vs-abd}

Abducible predicates are similar to open predicates, as their
extension is not specified in the program.  However, open predicates
can be given arbitrarily complex definitions (including all sorts of
cycles, possibly involving non-open predicates), while abducible
predicates are always defined by sets of ground facts (cf.\ the clause
$E\subseteq\abducibles(T,A)$ in Definition~\ref{def:gen-st-mod}).

Therefore, in order to relate the two approaches, we need a
representation lemma showing that a similar restriction can be posed
on the completions of open predicates, without loss of generality.

\begin{lemma}					\label{lem:repr}
If $M$ is a stable model of $P'\in\comp(P,F,O)$, then $M$ is a
stable model of some $P''\in\comp(P,F,O)$ such that $P''\setminus P$
is a set of ground facts.
\end{lemma}

\begin{proof}(Sketch)
Let $P''=P\cup\{p(\vec t)\in M \mid p\in O\}$.  It can be verified
that $M$ is also a stable model of $P''$.  Moreover, $P''\setminus P$
is a set of ground facts by definition.
\end{proof}

A rule $r = H\leftarrow B$ could also be abduced by introducing a new
abducible $n_r$ (that plays the role of $r$'s name), and inserting
$H\leftarrow B,n_r$ in the program (cf.\ \cite{Po88}).  Clearly, if
arbitrary rules are to be abduced (as it happens in open programs),
then this method leads necessarily to an infinite abduction framework
(more precisely, both the program and the set of abducibles are
infinite).  Lemma~\ref{lem:repr} improves rule abduction by showing
that in fact one needs not add any new symbols and rules, provided
that no constraint is posed on the space of abducible rules.

Now the basic correspondence between abduction and open programs
follows easily.

\begin{theorem}					\label{thm:abd&op}
$M$ is a generalized stable model of an abductive framework
$\tup{T,A}$ iff $M$ is a stable model of some
$P\in\comp(T,\emptyset,A)$.
\end{theorem}


The relationships between abductive frameworks and open programs
suggest two natural extensions of abductive reasoning under the
generalized stable model semantics.  First note that the set $F$ of
possible function symbols for the completions is empty in
Theorem~\ref{thm:abd&op}.  By allowing a nonempty $F$, the generalized
stable model semantics can be given the ability of assuming the
existence of new individuals and functions (as in \cite{DDS98} and
\cite{Sh89}).  Typically, in the framework of reasoning about action
and change, this feature is applied to explain a sequence of events
with unknown actions.  Abductive frameworks can be extended
accordingly, by a straightforward generalization of
$\abducibles(\cdot,\cdot)$.

\begin{definition}
For each abduction framework $\tup{T,A}$, let $\abducibles^o(T,A)$ be
a set of ground atoms $p(t_1,\ldots,t_n)$ such that $p\in A$ and $t_i$
is a term built from the function and constant symbols of $T$, plus a
denumerable set $Sk$ of (skolem) constants not occurring in $T$.
$\abducibles^o(T,A)$ will be called a set of \emph{open abducibles} of
the abduction framework $\tup{T,A}$ (w.r.t.\ the set of skolem
constants $Sk$).
\end{definition}

Clearly, the choice of $Sk$ is irrelevant, as long as it does not
intersect the vocabulary of $T$.

\begin{definition}				\label{def:open-abd}
An \emph{open generalized stable model} of an abduction framework
$\tup{T,A}$ is a stable model of $T\cup E$, for some
$E\subseteq\abducibles^o(T,A)$.
\end{definition}

By analogy with the previous theorem we have:

\begin{proposition}				\label{thm:abd&op2}
$M$ is an open generalized stable model of an abductive framework
$\tup{T,A}$ iff $M$ is a stable model of some $P\in\comp(T,Sk,A)$,
where $Sk$ is the set of skolem constants adopted in
$\abducibles^o(T,A)$.
\end{proposition}

The second extension concerns inference modalities.  Mixed inference
of type I suggests a new form of abduction that can be understood by
analogy with the two main forms of diagnosis, based on satisfaction
and entailment, respectively.  More precisely, diagnostic reasoning
can be carried out either by looking for a model of the domain theory
$T$ where the observations $Q$ are \emph{satisfied}, or by finding a
set of explanations $E$ such that $T\cup E$ \emph{entails} $Q$.  The
two kinds of diagnosis can be regarded as the analogues of standard
generalized stable model computation (which is equivalent to credulous
open reasoning, by Theorem~\ref{thm:abd&op}) and mixed open inference
of type I, respectively.  The latter can be reformulated for abductive
frameworks as follows.

\begin{definition}				\label{def:gen-skept-cons}
A sentence $Q$ is a \emph{generalized skeptical consequence} of an
abductive framework $\tup{T,A}$ if there exists
$E\subseteq\abducibles(T,A)$ such that $Q$ holds in \emph{all} the
stable models of $T\cup E$.

Similarly, $Q$ is an \emph{open generalized skeptical consequence} of
an abductive framework $\tup{T,A}$ if there exists
$E\subseteq\abducibles^o(T,A)$ such that $Q$ holds in \emph{all} the
stable models of $T\cup E$.
\end{definition}


Open programs improve standard abduction in one more respect: open
predicates can be \emph{partially specified} (cf.\ \cite{Bo-lpnmr01},
Example~14).  The same effect can be obtained by adding one more
abducible predicate for each partially defined predicate (cf.\
\cite{KKT92}).  A direct approach (such as the open framework) that
introduces no auxiliary symbols may be considered more elegant.

	\section{Open programs and ASP}
	\label{sec:op-vs-asp}

Open programs obviously generalize answer set programming.  If
$F=O=\emptyset$, then $\comp(P,F,O)=\{P\}$, and the four entailment
relations collapse to the standard credulous and skeptical entailment
of the stable model semantics.  

\begin{proposition}				\label{pro:op&asp}
Let $P$ be a normal logic program.  A sentence $Q$ is a credulous
(resp.\ skeptical) consequence of $P$ iff
$\tup{P,\emptyset,\emptyset}\crd Q$ (resp.\
$\tup{P,\emptyset,\emptyset}\skp Q$).  Moreover, if $P$ is consistent,
then $\tup{P,\emptyset,\emptyset}\crd Q$ is equivalent to
$\tup{P,\emptyset,\emptyset}\msc Q$\,, and
$\tup{P,\emptyset,\emptyset}\skp Q$ is equivalent to
$\tup{P,\emptyset,\emptyset}\mcs Q$.
\end{proposition}
In the rest of this section, we focus
on the opposite embedding.

Given Lemma~\ref{lem:repr}, embedding open programs into ASP may seem
trivial.  Apparently, the possible extensions of open predicates could
all be generated by suitable cyclic definitions, by analogy with the
embedding of generalized stable model semantics into (standard) stable
model semantics \cite{SI91}.  However, handling the different Herbrand
domains of the completions is not so trivial.

\begin{example}					\label{ex:naive-tran}
Let $P=\{p(a),\ q\leftarrow\neg p(X)\}$ and $F=\{b\}$.  Suppose that
$p\not\in O$ and $q\not\in O$.  The stable models of the completions
$P'\in\comp(P,F,O)$ satisfy $q$ iff $b$ occurs in $P'$.  As a
consequence, neither $q$ nor $\neg q$ are skeptical open consequences
of $\tup{P,F,O}$.  Now consider two naive attempts at embedding
$\tup{P,F,O}$ into ASP:
\begin{eqnarray*}
P_1 &=& P\cup \{p(a)\leftarrow\neg\bar p(a)\}\cup\{\bar
	p(a)\leftarrow\neg p(a)\}
\\
P_2 &=& P_1 \cup \{p(b)\leftarrow\neg\bar p(b)\}\cup\{\bar
	p(b)\leftarrow\neg p(b)\}
\end{eqnarray*}
(where $\bar p$ is a new predicate symbol).  Note that $\neg q$ is a
skeptical consequence of $P_1$ while $q$ is a skeptical consequence of
$P_2$, so none of the two programs is sound with respect to open
inference.
\end{example}

A faithful (partial) embedding of open programs into ASP can be
obtained by modeling the Herbrand domains explicitly (a device used
also for modeling quantification in Kripke structures where different
worlds have different domains).

\begin{definition}
Let $\tup{P,F,O}$ be an open program.  Define the corresponding normal
logic program $\Pi(P,F,O)$ as follows.  For each predicate symbol
$p\in O$, introduce a new distinct predicate symbol $\bar
p$. Moreover, let $U$, $S$ and $\bar S$ be three new predicate symbols
distinct from the symbols $\bar p$.  $\Pi(P,F,O)$ consists of the
following rules, for all rules $H\leftarrow Body$ in $P$, for all
$n$-ary function symbols $f$ occurring in $P$ or $F$, and for all
$p\in O$:
\[
\begin{array}{lp{8em}}
H\leftarrow Body, U(x_1),\ldots,U(x_n) &
where the $x_i$s are the variables of $H\leftarrow Body$
\\
U(f(x_1,\ldots,x_n)) \leftarrow \lefteqn{ U(x_1),\ldots,U(x_n),S(f) }
\\
\\
S(f) & if $f\not\in F$
\\
S(f) \leftarrow \neg \bar S(f) & if $f\in F$
\\
\bar S(f) \leftarrow \neg S(f) & if $f\in F$
\\
\\
p(x_1,\ldots,x_n) \leftarrow \neg \bar p(x_1,\ldots,x_n),
	\lefteqn{ U(x_1),\ldots,U(x_n) }
\\
\bar p(x_1,\ldots,x_n) \leftarrow \neg p(x_1,\ldots,x_n),
	\lefteqn{ U(x_1),\ldots,U(x_n) }
\end{array}
\]
\end{definition}

Intuitively, $S$ and $\bar S$ select a vocabulary from $P$ and $F$,
$U$ captures the corresponding ground terms (i.e., the Herbrand domain
of some completion), and the last rules generate an arbitrary set of
ground open atoms.

\begin{remark}
Strictly speaking, in the above definition the atoms $S(f)$ and $\bar
S(f)$ should be replaced by $S(\hat f)$ and $\bar S(\hat f)$, where
$\hat f$ is a \emph{name} for $f$.  For simplicity, here we use a
Prolog-like, relaxed syntax.
\end{remark}

\begin{example}					\label{ex:Pi}
Let $P$ and $F$ be as in Example~\ref{ex:naive-tran}, let $O=\{r\}$.
Then $\Pi(P,F,O)$ consists of the following rules:
\[
\begin{array}{l}
p(a)\\
q\leftarrow\neg p(X), U(X) \\
U(a) \leftarrow S(a) \\
U(b) \leftarrow S(b) \\
S(a)\\
S(b) \leftarrow \neg \bar S(b) \\
\bar S(b) \leftarrow \neg S(b) \\
r(X) \leftarrow \neg \bar r(X), U(X)\\
\bar r(X) \leftarrow \neg r(X), U(X)\,.
\end{array}
\]
\end{example}

The following theorem proves that the above embedding is correct.
Let $M|_{P,F,O}=\{\, p(t_1,\ldots,t_n)\in M\mid p \mbox{ occurs in
$\tup{P,F,O}$ and } U(t_i)\in M\ (1\leq i\leq n) \,\}.$

\begin{theorem}
$M$ is a stable model of $\Pi(P,F,O)$ iff there exist
$P'\in\comp(P,F,O)$ and a stable model $M'$ of $P'$ such that
$M|_{P,F,O}=M'$.
\end{theorem}

\begin{corollary}					\label{cor:op&asp}
For all sentences $Q$ with no occurrences of the new predicates $\bar
p$ ($p\in O$), $U$, $S$ and $\bar S$,
\begin{enumerate}
\item $\tup{P,F,O}\crd Q$ iff $Q$ is a credulous consequence of $\Pi(P,F,O)$.
\item $\tup{P,F,O}\skp Q$ iff $Q$ is a skeptical consequence of $\Pi(P,F,O)$.
\end{enumerate}
\end{corollary}

Note that $\Pi(P,F,O)$ may contain function symbols (then its
consequences may be undecidable).  If either $F$ or $O$ are infinite
then $\Pi(P,F,O)$ is infinite, otherwise $\Pi(P,F,O)$ is finite and it
can be computed in polynomial time (w.r.t.\ $|P\cup F\cup O|$).  The
following corollary immediately follows.

\begin{corollary}				\label{cor:complxity1}
Suppose that $P$, $F$ and $O$ are finite and contain only 0-ary
function symbols.
\begin{enumerate}
\item If $P$ is ground, then credulous and skeptical open inferences
are NP-complete and coNP-complete, respectively.
\item If $P$ is not ground, then credulous and skeptical open
inferences are NEXPTIME-complete and coNEXPTIME-complete,
respectively.
\end{enumerate}
\end{corollary}

Corollary~\ref{cor:op&asp} and Corollary~\ref{cor:complxity1} cannot
be easily extended to mixed open inference.  The above translation
technique (that basically derives from Lemma~\ref{lem:repr}) captures
faithfully the credulous consequences of the completions, but
apparently it does not scale to their skeptical consequences.
Currently, we do not know whether mixed inference can be polynomially
embedded into ASP.

The results of the previous section can be combined with the above
translation to obtain an embedding of abduction into ASP.

\begin{corollary}				\label{cor:abd2asp}
For all interpretations $M$,
\begin{enumerate}
\item
$M$ is a generalized stable model of an abduction framework
$\tup{T,A}$ iff $M=M'|_{T,\emptyset,A}$\,, for some stable model $M'$
of $\Pi(T,\emptyset,A)$.
\item
$M$ is an open generalized stable model of an abduction framework
$\tup{T,A}$ iff $M=M'|_{T,Sk,A}$\,, for some stable model $M'$
of $\Pi(T,Sk,A)$ (where $Sk$ is the set of skolem constants adopted in
$\abducibles^o(T,A)$).
\end{enumerate}
\end{corollary}

The translation $\Pi(T,\emptyset,A)$ is slightly more redundant than
the translation introduced in \cite{SI91}, but a trivial unfolding
process yields an equivalent embedding.

\begin{example}
Consider the abduction framework $\tup{P,O}$, where $P$ and $O$ are
specified as in Example~\ref{ex:Pi}.  $\Pi(P,\emptyset,O)$ is
\[
\begin{array}{l}
p(a)\\
q\leftarrow\neg p(X), U(X) \\
U(a) \leftarrow S(a) \\
S(a)\\
r(X) \leftarrow \neg \bar r(X), U(X)\\
\bar r(X) \leftarrow \neg r(X), U(X)\,.
\end{array}
\]
Unfolding yields the (expected) simplified program
\[
\begin{array}{l}
p(a)\\
q\leftarrow\neg p(a) \\
r(a) \leftarrow \neg \bar r(a) \\
\bar r(a) \leftarrow \neg r(a) \,.
\end{array}
\]
\end{example}

The translation $\Pi(T,Sk,A)$ extends the approach by \cite{SI91} with
the ability of abducing new individuals.  Unfortunately, $\Pi(T,Sk,A)$
is infinite if $Sk$ is infinite.  Therefore, in the absence of any
upper bound on the domain's cardinality, the existing credulous
engines for ASP can only be applied to a finite approximation
$\Pi(T,Sk',A)$ of the original problem, where $Sk'$ is a finite subset
of $Sk$.  Clearly, this approximation yields all and only the
abductions that postulate the existence of at most $|Sk'|$
individuals, besides those explicitly mentioned in the domain
knowledge.

\begin{example}
Let
\begin{eqnarray*}
T & = & \{\ p(a),\ q \leftarrow r(X), \neg p(X)\ \} \\
A & = & \{\, r \,\} \,.
\end{eqnarray*}
No generalized stable model of $\tup{T,A}$ satisfies $q$, whereas if
$Sk = \{ s_0, s_1, \ldots, s_i, \ldots \}$, then  $\tup{T,A}$ has an
open generalized stable model $$\{ p(a), q, \} \cup \{ r(s_i) \mid i\in
I \}$$ for each set of natural numbers $I$.  Only one individual needs
to be abduced in order to explain $q$.  Accordingly, the program
$\Pi(T,\{s_0\},A)$ consisting of the rules
\[
\begin{array}{l}
p(a)\\
q\leftarrow r(X),\neg p(X), U(X) \\
U(a) \leftarrow S(a) \\
U(s_0) \leftarrow S(s_0) \\
S(a)\\
S(s_0) \leftarrow \neg \bar S(s_0) \\
\bar S(s_0) \leftarrow \neg S(s_0) \\
r(X) \leftarrow \neg \bar r(X), U(X)\\
\bar r(X) \leftarrow \neg r(X), U(X)
\end{array}
\]
has a stable model $\{ p(a), q, r(s_0) \}$, that ``explains'' $q$ with
$\{ r(s_0) \}$.  In this case, the bound on the open domain does not
cause any significant loss of information, in the sense that all the
alternative minimal explanations are isomorphic to $\{ r(s_0) \}$.
\end{example}

	\section{Summary and Conclusions}
	\label{sec:conc}

Open programs constitute a simple unifying framework that generalizes
both abduction frameworks (under the generalized stable model
semantics) and answer set programming, as shown by
Theorem~\ref{thm:abd&op} and Proposition~\ref{pro:op&asp}.  In
particular, any abduction framework $\tup{T,A}$ corresponds to the
open program $\tup{T,\emptyset,A}$, while each normal logic program
$P$ is captured by the degenerate open program
$\tup{P,\emptyset,\emptyset}$.

The reduction of the generalized stable model semantics to open
inference highlights similarities and differences between the two
frameworks.  In particular, it shows that the only real difference
lies in the treatment of the domains, that are fixed in abduction
frameworks, and variable in the open framework (cf.\ the discussion
after Theorem~\ref{thm:abd&op}).  On the contrary, the different kinds
of predicate specifications considered by the two frameworks (ground
facts vs.\ arbitrary definitions) have no influence, as they select
the same set of models (Lemma~\ref{lem:repr}).  This result implies
also that in the absence of constraints on the space of abducible
rules, no new abducible predicates and rules need to be introduced in
order to abduce complex sentences (as opposed to the standard
formulation \cite{Po88}).

We have applied the open framework to give the generalized stable
model semantics the ability of abducing new individuals
(Definition~\ref{def:open-abd}).  Then we extended the embedding of
abduction into ASP accordingly, using the more general result for open
programs (Corollary~\ref{cor:abd2asp}).  As a result, ASP engines such
as Smodels can be applied to abductive reasoning over \emph{bounded}
open domains (cf.\ the discussion after Corollary~\ref{cor:abd2asp}).
Because of the bound on the domain size, ASP can only approximate
abduction with unbounded open domains.  We are currently generalizing
\emph{finitary programs} \cite{Bo-IJCAI01} to achieve semidecidable,
\emph{exact} abduction with unbounded open domains.

Note, however, that the bound on domain size is not always a
restriction.  In some cases, the bound may be part of the domain
knowledge.  The existing calculi that support open domains do not
express nor reason about such restrictions \cite{DDS98,Sh89}.

The open framework suggests a new form of abduction analogous to
\emph{diagnosis-as-entailment}, called \emph{generalized skeptical
consequences} (Definition~\ref{def:gen-skept-cons}).

Our results improve the understanding of open inference.  From the
embedding into ASP we derived the complexity of skeptical and
credulous open inference in the finite case, and in the absence of
proper functions (Corollary~\ref{cor:complxity1}).  In the other
cases, the translation may be infinite or contain function symbols,
and the only known automated deduction method is the skeptical open
resolution calculus introduced in \cite{Bo-lpnmr01}.

This calculus derives sentences that hold no matter how the open
program is completed.  It may be interesting to investigate the use of
such skeptical inferences to pre-compute lemmata, in order to speed-up
the computation of explanations.

Moreover, the skeptical open calculus is nonground, and may return
nonground answers.  This may be a starting point for nonground
abduction under the open generalized stable model semantics.

On the other hand, Theorem~\ref{thm:abd&op} and
Corollary~\ref{cor:op&asp} open the way to the application of
abduction procedures and ASP engines to open inference problems, at
least under some cardinality and arity restrictions.  In particular,
abduction and ASP might be applied to static verification problems for
various kinds of incomplete specifications.

Currently it is not clear whether the mixed forms of open inference
can be embedded into ASP and abduction frameworks.  We conjecture that
no polynomial reduction based on normal logic programs exists.  It
should be verified whether the calculus for mixed inference
illustrated in \cite{Bo-lpnmr01} can be immediately applied to the
generalized skeptical consequences introduced in
Definition~\ref{def:gen-skept-cons}.

 	\subsubsection*{Acknowledgements}

The author is grateful to the anonymous referees for their comments,
that led to significant improvements in the paper.

The work reported in this paper was partially supported by the
European Community within the Fifth (EC) Framework Programme
under contract IST-1999-11791 -- FASTER project.

\end{document}